\documentclass[fleqn,10pt]{wlscirep}
\usepackage[utf8]{inputenc}
\usepackage[T1]{fontenc}
\graphicspath{ {figures/} }
\usepackage{diagbox}
\usepackage{makecell}
\usepackage{comment}

\title{Untrained neural networks can demonstrate memorization-independent abstract reasoning}

\author[1,*]{Tomer Barak}
\author[1,2]{Yonatan Loewenstein}
\affil[1]{The Edmond and Lily Safra Center for Brain Sciences, The Hebrew University, Jerusalem, Israel}
\affil[2]{Department of Cognitive Sciences, The Federmann Center for the Study of Rationality, The Alexander Silberman Institute of Life Sciences, The Hebrew University, Jerusalem, Israel}

\affil[*]{tomer.barak@mail.huji.ac.il}

\begin{abstract} % Maximum 200 words.
The nature of abstract reasoning is a matter of debate. Modern artificial neural network (ANN) models, like large language models, demonstrate impressive success when tested on abstract reasoning problems. However, it has been argued that their success reflects some form of memorization of similar problems (data contamination) rather than a general-purpose abstract reasoning capability. This concern is supported by evidence of brittleness, and the requirement of extensive training. In our study, we explored whether abstract reasoning can be achieved using the toolbox of ANNs, without prior training. Specifically, we studied an ANN model in which the weights of a naive network are optimized during the solution of the problem, using the problem data itself, rather than any prior knowledge. We tested this modeling approach on visual reasoning problems and found that it performs relatively well. Crucially, this success does not rely on memorization of similar problems. We further suggest an explanation of how it works. Finally, as problem solving is performed by changing the ANN weights, we explored the connection between problem solving and the accumulation of knowledge in the ANNs.
\end{abstract}

\begin{document}

\flushbottom
\maketitle

\thispagestyle{empty}

\section*{Introduction}

The topic of this paper is abstract reasoning, sometimes referred to as “fluid intelligence” \cite{lohman_complex_2000}. Abstract reasoning is, broadly speaking, the ability to solve complex problems by identifying regularities and relations in the problem being solved and utilizing them for deducing the \cite{sternberg_component_1977,sternberg_components_1983}. It is often studied using intelligence tests that comprise word analogy tests (e.g., infer that the relationship between “cow” and “milk” is the same as between “chicken” and “egg”) and visual reasoning tests (e.g., Raven Progression Matrices) \cite{kaplan_psychological_2009,raven_manual_1998}. As artificial intelligence continues to advance, understanding the nature of abstract reasoning in both humans and machines is becoming a central question in cognitive science and AI research \cite{mitchell_how_2023}.

Abstract reasoning in artificial neural networks (ANNs) appears to be closely tied to training. While deep ANNs have shown impressive performance on various intelligence tests \cite{hersche_neuro-vector-symbolic_2023, santoro_simple_2017, barrett_measuring_2018, openai_gpt-4_2023, kim_few-shot_2020, webb_emergent_2023}, their success relied heavily on extensive prior training. Additionally, questions have been raised about the nature of this performance. There are indications that ANNs’ success may stem more from “contamination” – exposure to similar questions in their training data – rather than from genuine abstract reasoning \cite{mccoy_embers_2023, biever_chatgpt_2023}. This dependency on specific training data is further emphasized by findings that minor changes in problem phrasing, which do not affect human performance, can render problems unsolvable for ANNs \cite{azulay_why_2019, mitchell_how_2023}. Thus, while ANNs may exhibit some analogical reasoning capabilities, it is disputed that these are based on pattern matching or memorization rather than on general intelligence comparable to that of humans.

In this work, we investigated whether ANN tools commonly used in machine learning are capable of demonstrating general abstract reasoning. Specifically, we asked if these networks could solve intelligence test problems with novel inputs, relying only on the information provided by the specific problem at hand, without drawing on prior memorization.

Certain intelligence tests, by their nature, require some level of prior knowledge. For instance, a human unfamiliar with English or the relationship between ``cow'' and ``milk'' would struggle to relate ``chicken'' to ``egg'' in an analogy test. Consequently, general intelligence in humans is often assessed using \textit{visual} reasoning tests, utilizing abstract shapes like squares and triangles to minimize the influence of language or cultural knowledge. Thus, to evaluate the abstract reasoning of ANN models, we employed visual abstract reasoning tests. These visual reasoning tests require identifying relations in a sequence of stimuli, a skill common to many intelligence tests \cite{sternberg_component_1977, sternberg_components_1983, lohman_complex_2000, siebers_computer_2015}.

For our network models, we used Relation Networks (RNs) \cite{sung_learning_2018}, as members of this class of models were shown to be capable of identifying abstract relations and solving intelligence tests after extensive training \cite{hill_learning_2018, barrett_measuring_2018}. Notably, RNs were shown to successfully solve word-analogy problems without specific training on those problems, but with specific training on relevant relationships \cite{lu_emergence_2019,  lu_probabilistic_2022}. In contrast to these previous studies, our focus was on the ability of ``naive'' RNs, who were not exposed to \textit{any} pre-training, to identify relations in visual reasoning tests and use them for solving the tests.

The structure of this paper is as follows: We begin by introducing the visual reasoning tests and the network models employed in our study. Next, we present the model’s performance, showcasing its ability to solve non-trivial problems. We then analyze the mechanisms underlying this performance. Finally, as the model’s problem-solving involves an optimization process that modifies network parameters in a manner similar to learning, we examine the relationship between the model’s problem-solving capabilities and the networks’ accumulation of knowledge.

\section*{Results}

\subsection*{Sequential visual reasoning tests}

\begin{figure}[h!]
\begin{center}
\centerline{\includegraphics[width=\linewidth]{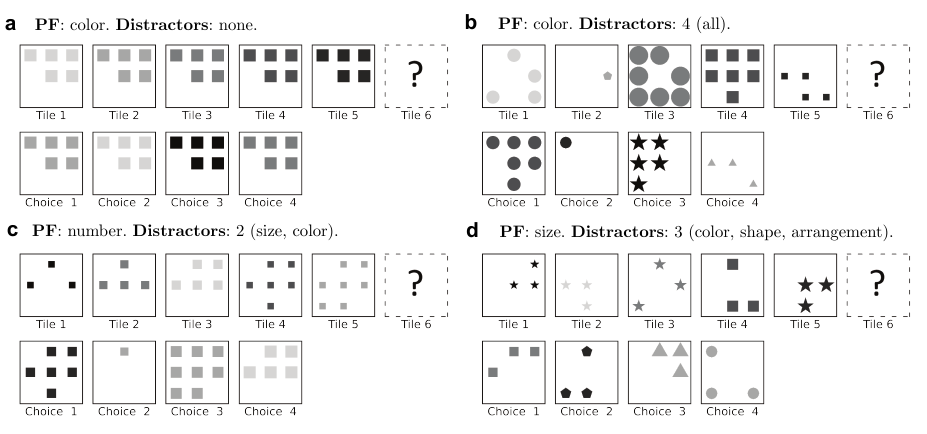}}
\caption{\textbf{Visual reasoning problems.} The problems are characterized by the \emph{Predictive Features (PF)} that can be the color (\textbf{a-b}), number (\textbf{c}), or size (\textbf{d}) of the abstract shapes. The values of the predictive features linearly increase along the sequence. The rest of the features (non-predictive) are either constant or random. We refer to the random features as \emph{Distractors}, and their number determines the problem difficulty. Note: the shapes' type and arrangement are always non-predictive, and can either be constant or distracting. The correct choices in this figure are all $3$.}
\label{fig:sequences_examples}
\end{center}
\end{figure}

We constructed a set of artificial problems in which the task is to evaluate the consistency of an image with a sequence of its preceding images (Fig. \ref{fig:sequences_examples}). Each problem comprises $5$ gray-scale images and $4$ optional-choice images. The images, $224 \times 224$ pixels each, are composed of identical abstract objects and differ along several dimensions: the shape of the objects, their size, their color, their number, and their arrangement. By construction, one of these features changes predictably over the $5$ images. Formally, an image is characterized by a low-dimensional vector of features, $\mathbf{f}_j$, where $f^i_j$ denotes the value of feature $i$ in image $j$. An image in pixel space, $\mathbf{x}_j$, is constructed according to its characterizing features by a generative function $\mathbf{x}_j=\mathbf{G}\left(\mathbf{f}_j\right)$. One of the features $f^p$ changes predictably along the sequence according to a simple deterministic rule $f_{j+1}^p=U(f_j^p)$ while the other features are either constant over the images or change randomly (values are i.i.d). Considering the optional-choice images, the predictable rule is followed in only one of them, and the task is to select this image. The other features are either constant in all $9$ images ($5$ of sequence and $4$ of optional choices) or change randomly (see Methods). We refer to a problem's predictably changing feature as the problem's Predictive Feature (PF) and to the randomly changing features as distracting features or \emph{distractors}. Intuitively, the number of distractors is a measure of a problem's difficulty.

\subsection*{The computational task}

Each image is characterized by a small number of features, of which one changes predictably. The challenge is to simultaneously identify the features and the rule that relates the features of the different images. Relation Networks \cite{sung_learning_2018} do exactly that. Taking a set of stimuli, they learn two functions: an encoder function $Z_\phi\left(\mathbf{x}\right)$ that extracts relevant feature(s) from the stimulus, that is, a low-dimensional representation of the stimuli $\mathbf{x}$  and a relation module $R_\theta\left(Z_\phi\left(\mathbf{x}_i\right), Z_\phi\left(\mathbf{x}_j\right)\right)$ that characterizes the relationship between the features of pairs of stimuli $\mathbf{x_i}$ and $\mathbf{x_j}$. In practice, the encoder and the relation modules are functions (typically networks) whose parameters ($\phi$ and $\theta$, respectively) are learned from examples. It should be noted that, to some extent, the complexities of the encoder and the relation module are interchangeable. The reason is that a sufficiently-complex relation module can incorporate the feature extraction. Similarly, a sufficiently complex encoder can operate on the extracted features as to simplify the relation between them. For example, any monotonous relation between the features is also a linear relation between a (nonlinear) transformation of the features.

Previous studies have shown that with sufficient examples, relational networks can learn to extract the relevant features and their relations at a level sufficient for solving intelligence tests \cite{barrett_measuring_2018, hill_learning_2018, lu_probabilistic_2022}. The challenge here is to perform a similar task without any pre-training. To do so, we defined the following loss function on a sequences of $5$ images:

\begin{equation}
\label{eq:Loss}
    \mathcal{L}(\theta,\phi)= \frac{1}{4}\sum_{i=1}^{4} \left[ R_\theta\left(Z_\phi\left(\mathbf{x}_i\right), Z_\phi\left(\mathbf{x}_{i+1}\right)\right)\right]
\end{equation}
$\mathbf{x}_i$ is the $i^\text{th}$ image in the sequence, the encoder $Z_\phi: \mathbb{R}^{224\times224}\rightarrow\mathbb{R}^n$ is a function that takes $224\times224$ pixel images to an $n$ dimensional latent space and the relation module $R_\theta: \mathbb{R}^{2n}\rightarrow \mathbb{R}^+$ takes two consecutive latent variables, each of dimension $n$, and outputs a positive $1D$ relation score. 

This loss is minimized for a relation function $R_\theta\left(Z_\phi\left(\mathbf{x}_i\right), Z_\phi\left(\mathbf{x}_j\right)\right)$ that outputs a minimal relation score for consecutive sequence images ($j=i+1$), requiring the identification of the regularity that characterizes these consecutive images. We updated the networks' weights $\theta$ and $\phi$ with 10 optimization steps over the loss $\mathcal{L}\left(\theta,\phi\right)$ using the RMSprop optimizer \cite{dauphin_rmsprop_2015} (learning rate of $10^{-5}$, the rest of the parameters are set to PyTorch \cite{paszke_pytorch_2019} default). Eventually, after optimization, we evaluated the consistency of each choice image with the sequence based on their relation value $R$ when they were placed as the sixth sequence image $R_\theta\left(Z_\phi\left(\mathbf{x}_5\right), Z_\phi\left(\ \cdot\ \right)\right)$ and selected the choice image with the lowest relation value as the answer.

To clarify, in these settings, the model does not need to learn the features and their relation in the generative sense to solve a test successfully. Instead, it is enough to find image representations and rules that are \emph{sufficiently} correlated with a problem's predictive feature for selecting the most consistent image out of four options.

\subsection*{Vanilla model performance}

The success of the model would depend on the specific choice of $R$ and $Z$ (their network structure), as they can be inductively biased towards certain types of features and rules. In our vanilla model, the encoder $Z$ was a small CNN from input space to a 1D latent neuron, composed of 3 convolutional layers followed by 5 fully-connected (FC) layers with a single output neuron (see Methods and Supplementary Information Fig. S1). For the relation module, we used a simple function that asserts a linearly changing relation between the latent variables,
\begin{equation} R_\theta\left(Z_\phi\left(\mathbf{x}_i\right), Z_\phi\left(\mathbf{x}_j\right)\right)=\left( Z_\phi\left(\mathbf{x}_i\right)  - Z_\phi\left(\mathbf{x}_{j}\right) + \theta \right)^2
\end{equation}
where $\theta$ is a trainable constant that does not depend on $Z$.

\begin{figure}[h!]
% \vskip 0.2in
\begin{center}
\centerline{\includegraphics[width=\linewidth]{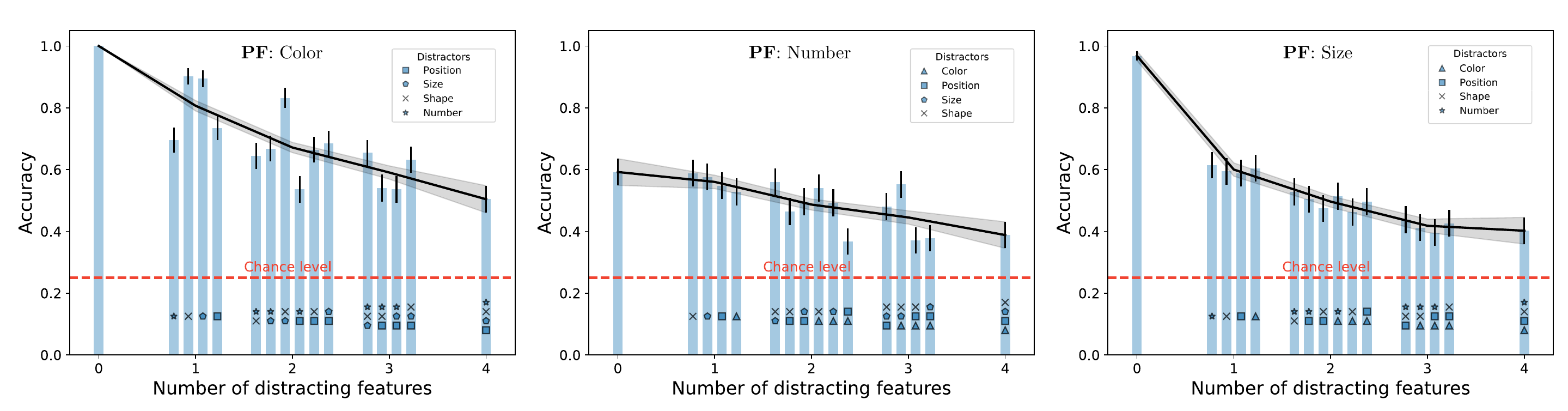}}
% \vskip -0.05in
\caption{\textbf{Vanilla model performance.} The performance of naive ANNs on the three Predictive Features (PFs): Color (left), Number (center), and Size (right). For each predictive feature, we tested the networks over 16 test conditions where the predictive feature was linearly changing along the sequence, and the non-predictive features were either distractors (marked according to the legend) or constant (not marked). Each test condition included $500$ randomly generated problems. Error bars are 95\% Confidence Intervals (CI). The black line and its shade are the average accuracy per difficulty and the corresponding 95\% CI. The dashed line denotes the chance level of problems with four choice images ($0.25$).}
\label{fig:MCPC_mon}
\end{center}
% \vskip -0.3in
\end{figure}

We evaluated the performance of the vanilla model on the different tests, in which the predictive feature's values increased linearly, and found that it performed substantially better than chance ($0.25$) in almost all tasks and all levels of difficulty (Fig. \ref{fig:MCPC_mon}). Without distractors, its performance on some tasks was close to perfect. We also found that performance decreased with the number of distractors, verifying that the number of distractors is a good measure of the task's difficulty. All these results were obtained using networks that we randomly initialized before each problem, thus demonstrating that ANNs can perform abstract reasoning that does not depend on memorization. Averages over all conditions, the model's performance was $\mathbf{0.58\pm0.01}$. From this point in the paper, we use this global performance measure for comparisons (see Methods; complete performance results are in the Supplementary Information).

\subsection*{Determinants for success}

Our model consists of two main components: the encoder $Z$ and the relation module $R$. The parameters of both were changed in the direction of minimizing the loss function on the images of each problem, a process that we will refer to as optimization. To study the relative contribution of these components to problem-solving, we studied the model's performance when the parameters of only one of these components, either $Z$ or $R$, were optimized. We found that optimizing the encoder was essential: when the parameters of the encoder $Z$ remained unchanged, the model's performance, averaged over all conditions, was close to the chance level, $0.30\pm0.01$ (Fig. S2). By contrast, using random parameters for the relation module $R$ had no significant effect on performance, resulting in an average performance of $0.58\pm0.01$ (Fig. S3), which is not significantly different from that of the vanilla model. These results motivated us further to study the role of the encoder in the task.

\subsubsection*{The encoder}

The encoder is an 8-layer network with $3$ convolutional layers followed by $5$ Fully-Connected (FC) layers. Removing the convolutional layers and connecting the FC layers directly to the inputs impaired the average performance of the model, reducing its performance to $0.48\pm0.01$ (Fig. S4), indicating that the convolutional layers are important for performance. In the vanilla model, the parameters of both the convolutional layers and the FC layers are optimized in the direction of minimizing the loss function. However, it turns out that the optimization of the parameters of the convolutional layers does not contribute to the performance. The average performance when the weights of the convolutional layers remained random, $0.57\pm0.01$, was not significantly different than that of the vanilla model (Fig. S5). By contrast, keeping the FC network weights fixed at their randomly-initialized values during problem-solving was detrimental to the performance ($0.34\pm0.01$, Fig. S6). 

So far, we saw that freezing either the weights of the convolutional layers or the relation module at their initial random values does not impair performance. This insensitivity does not change when \emph{both} are frozen ($0.58\pm0.01$, Fig. S7). 

We conclude that the convolutional layers effectively operate as frozen feature extractors (features in the more general sense -- not necessarily the features used for constructing the images) while the parameters of the FC layers are optimized to solve the task.

\begin{figure}[h!]
% \vskip 0.2in
\begin{center}
\centerline{\includegraphics[width=\linewidth]{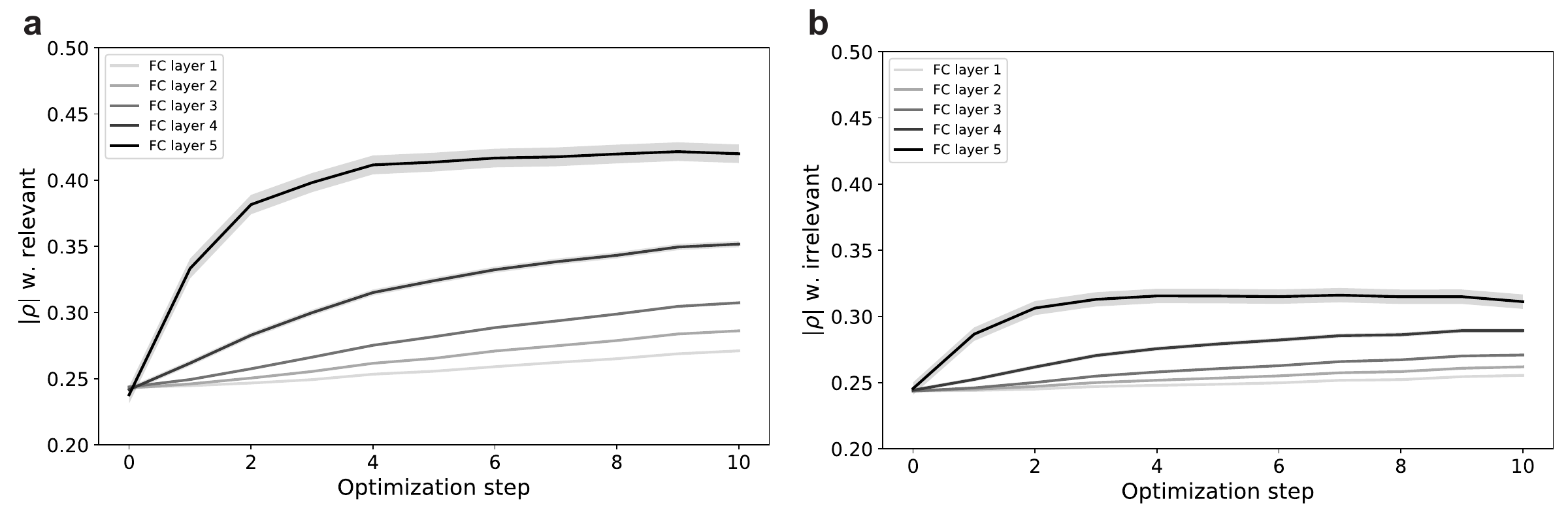}}
% \vskip -0.05in
\caption{\textbf{The encoder's FC layers feature correlations.} The average absolute correlations of encoders' FC layers with (\textbf{a}) the specific predictive feature of the problems they solved (either Color, Number, or Size), and (\textbf{b}) the other two non-predictive features (from either Color, Number, or Size). Error shades represent the 95\% CI, based on the standard error of the means. The calculation of the correlations is detailed in the Methods section.}
\label{fig:feat_corr}
\end{center}
% \vskip -0.3in
\end{figure}

To test how the FC layers contribute to this task, we note that the task could be perfectly solved if the encoder could learn to identify the inverse generative function of the problem images $G^{-1}(\mathbf{x})$ and use it to extract the underlying predictive feature $f^p$ and its rule $U(f^p)$. If this is done, we expect the optimization steps to increase the correlation of the encoder's output neuron with the predictive feature (but not with the distracting features). We tested this hypothesis in the vanilla model for all the predictive features. Indeed, the absolute Pearson correlation of the output neuron with the predictive feature (see Methods) increases with optimization steps, as depicted in Fig. \ref{fig:feat_corr}\textbf{a} (black). 

To better understand how such correlations emerge in the FC network, we also computed the absolute Pearson correlations of these features with the activities of all other neurons in the FC network (see Methods. Comprehensive results in Supplementary Fig. S8). These correlations, averaged over all neurons in a layer, are depicted in Fig. \ref{fig:feat_corr}\textbf{a}. We found that the correlations with the predictive feature increase with the layer depth. 

The higher the correlation of the output neuron with the predictive feature, the easier it is for the relation module to identify the regularity in the sequence of images. Along the same lines, we also expected the optimization process to decrease the correlation of the encoder output neuron with the other irrelevant features. This, however, is not the case. Considering the same features in problems in which they are not predictive features (either constant or distracting), we found that the correlation of the output neuron with these features also increases on average in the optimization process, albeit to a lesser extent (Fig. \ref{fig:feat_corr}\textbf{b}). Considering the correlations of these features with neurons in the hidden layers of the encoder, we found that the correlations with these irrelevant features also increased with the layer depth. 

\begin{figure}[h!]
% \vskip 0.2in
\begin{center}
\centerline{\includegraphics[width=0.5\linewidth]{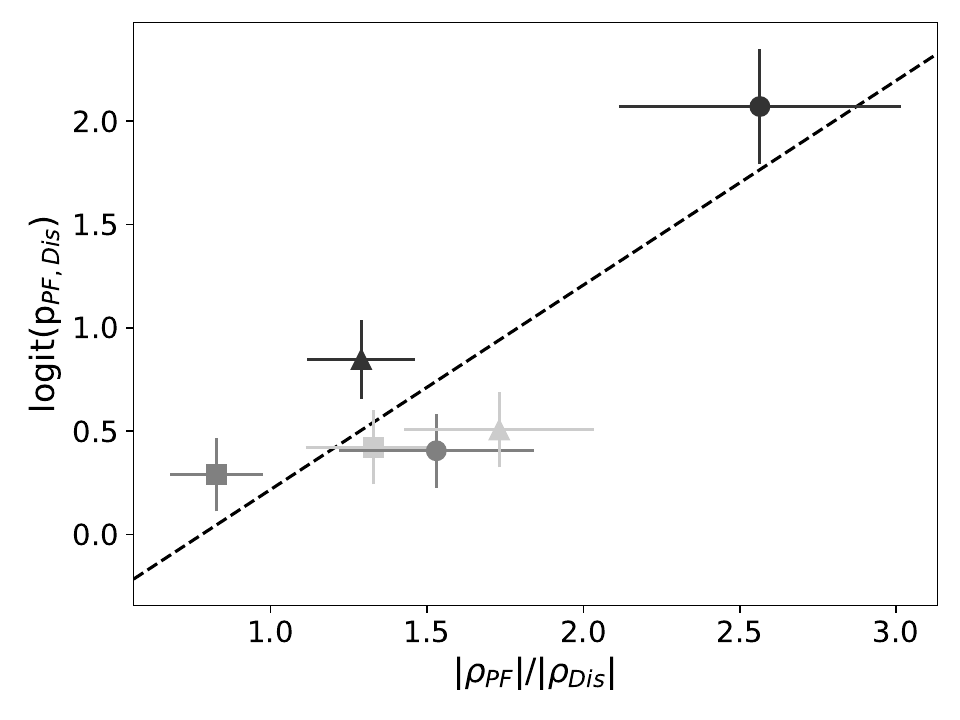}}
% \vskip -0.05in
\caption{\textbf{The effect of distractors on accuracy.} The figure depicts the relationship between the absolute correlation ratio with the relevant Predictive Feature ($|\rho_{PF}|$) and the Distracting feature $|\rho_{Dis}|$, and its consequential effect on networks' accuracy in problems of that predictive feature with the corresponding distracting feature ($p_{PF,Dis}$). The predictive features were either Color (Dark Gray), Number (Medium Gray), or Size (Light Gray). The distracting features were either Color (Square), Number (Triangle), or Size (Circle). Error bars represent the 95\% CI. The black dashed line depicts a linear regression analysis.}\label{fig:performance_vs_corr}
\end{center}
% \vskip -0.3in
\end{figure}

At the end of the optimization procedure, the output neuron of the encoder network is correlated with both the predictive feature and the irrelevant features (both distractors and constant). The stronger the correlation of the output neuron with the predictive feature, relative to its correlation with the distracting features, the better we expected the performance to be. To test this, we focused on the six problems in which the relevant feature was either color, number, or size, and there was one distracting feature, again: color, number, or size. For each of these problems, we computed the absolute Pearson correlations of the output neuron with the predictive and distracting features (taken from Fig. S8). We expected that performance in each of these problems would increase with the correlation with the relevant feature and decrease with the correlation with the irrelevant feature. Indeed, as depicted in Fig. \ref{fig:performance_vs_corr}, the logit of the performance ($\log\frac{p}{1-p}$ where the accuracies $p$ are taken from Fig. \ref{fig:MCPC_mon}) is correlated with the ratio of the absolute Pearson correlation of the output neuron with the predictive feature and the distracting feature (Wald t-test, p-value = 0.028).

\begin{figure}[h!]
% \vskip 0.2in
\begin{center}
\centerline{\includegraphics[width=\linewidth]{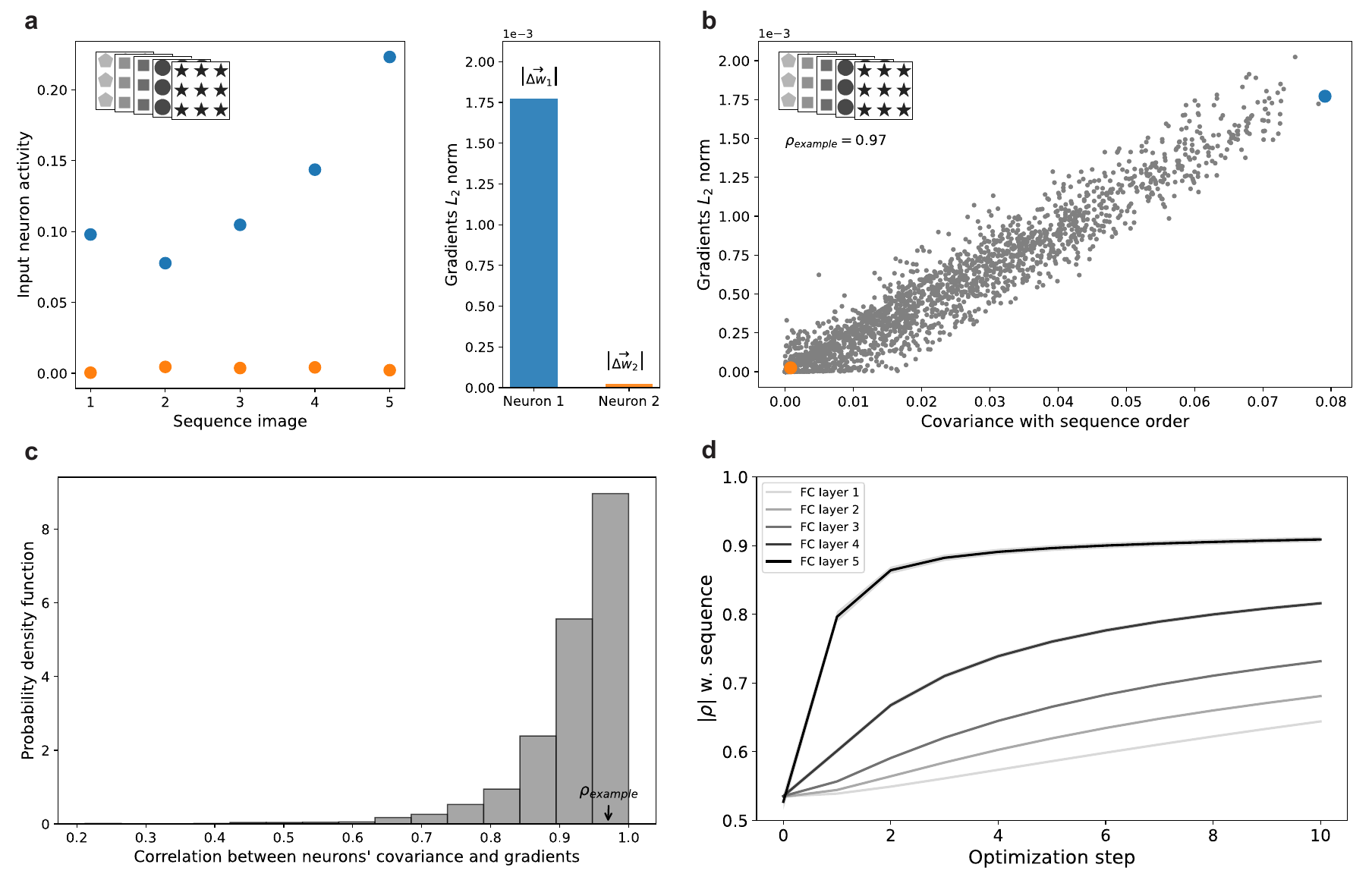}}
% \vskip -0.05in
\caption{\textbf{Problem-solving mechanism.} (\textbf{a}) Two example neurons' activity from the convolutional layers' output of a network (before optimization) when presented with the example problem of the inset. The blue neuron has a large covariance with the problem's image order, and the orange neuron has a small covariance with the order. The neurons' L2 gradient norms correlate with their respective image order covariances. \textbf{(b)} In this example network, the L2 gradient norms of the convolutional layers' output neurons are strongly correlated with their image-order covariances ($\rho_{\text{example}}=0.97$). The two example neurons presented in (a) are highlighted. \textbf{(c)} Distribution of the correlations between L2 gradient norms and images' sequence-order across all problems. (\textbf{d}) The absolute correlation of the encoder's FC layers with the sequence order during the optimization process. Error shades represent 95\% CI. Neurons' covariance and correlation calculations are explained in Methods.}
\label{fig:seq_corrs}
\end{center}
% \vskip -0.3in
\end{figure}

Next, we studied how the correlation with the features increases during optimization. The loss function ``seeks'' \emph{some} 1D predictable representation of the sequence of inputs. Considering the individual neurons at the output layer of the convolutional network part of the encoder, some co-vary with the sequence while others do not. Examples of two such neurons are depicted in Fig. \ref{fig:seq_corrs}\textbf{a} (left). As shown in Fig. \ref{fig:seq_corrs}\textbf{a-b} for a single problem, the optimization process makes larger changes to the synaptic weights from those neurons that co-vary strongly with the sequence order (e.g., blue in Fig. \ref{fig:seq_corrs}\textbf{a-b}) compared with low co-variance neurons (e.g., orange in Fig. \ref{fig:seq_corrs}\textbf{a-b}). This is the case across all problems (Fig. \ref{fig:seq_corrs}\textbf{c}). Consequently, the neurons in the encoder's FC layer become strongly correlated with the sequence order (Fig. \ref{fig:seq_corrs}\textbf{d}). As a result, the encoder amplifies the representation of those features that co-vary with the sequence order (independently of whether they are the predictive or irrelevant features).

Together, our results indicate that the ANN's ability to execute abstract reasoning without prior learning stems from two important properties: (1) The random convolutional layers extract features that are correlated with the relevant features. (2) The optimization process amplifies the response to those features that monotonically vary along the sequence. 

\subsubsection*{The relation module}
\begin{figure}[h!]
\begin{center}
\centerline{\includegraphics[width=\linewidth]{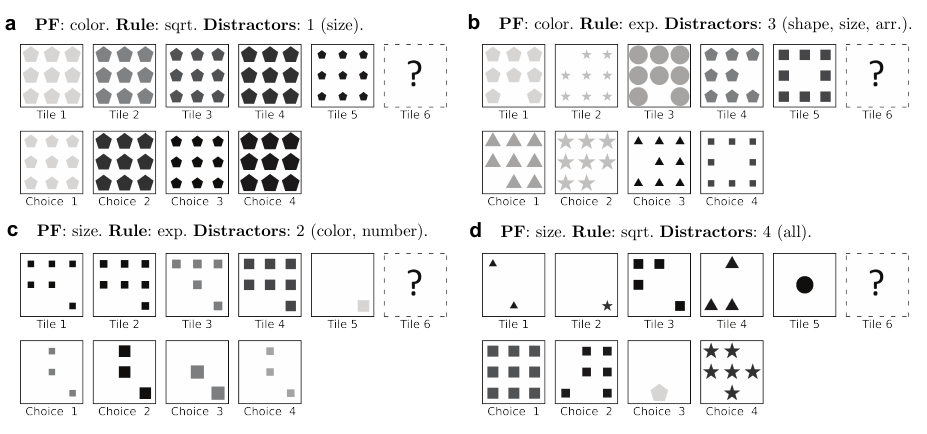}}
\caption{\textbf{Non linear rules.} The predictive features' values in these tests increased as a square root (\textbf{a} and \textbf{d}) or exponentially (\textbf{b} and \textbf{c}). The rest of the features (non-predictive) are either constant or random. \emph{The correct choices are all $3$}.}
\label{fig:sequences_examples_nonLinear}
\end{center}
\end{figure}

By construction, the vanilla model's relation module is simple, implicitly assuming that the features change linearly. Therefore, one may naively expect that identifying a non-linear change in the predictive feature will be more challenging. However, any monotonically changing rule can be mapped into a linearly changing rule with a sufficiently complex encoder. We, therefore, tested our vanilla model in problems in which the change in the feature was \emph{non-linear} (Fig. \ref{fig:sequences_examples_nonLinear}). We found that when the relevant feature was size, the performance for an exponential increase or a square root increase of this feature was comparable to that of a linear increase (Linear: $0.53\pm0.01$; Exp: $0.54\pm0.01$; Sqrt: $0.52\pm0.01$. Fig. S9-10 right). Similarly, when the relevant feature was color, the model achieved comparable performance to the linear case, although with higher variability: performance was better for an exponential increase and worse for a square root increase (Linear: $0.70\pm0.01$; Exp: $0.75\pm0.01$; Sqrt: $0.64\pm0.01$. Fig. S9-10 left). These results suggest that a relation module that assumes linear relationships can capture general monotonic relationships, substantially downsizing the hypothesis space of possible relationships. It would, however, be more difficult for the model to deal with non-monotonic rules. Indeed, when tested in problems where the predictive feature alternated between two of its values, the vanilla model performance was at a chance level ($0.24\pm0.01$. Fig. S11).

In the vanilla model, the relation module is simple and general, and the encoder that finds appropriate image representations carries most of the ``computational load''. However, we expect the complexity of the encoder and the complexity of the relation module to be interchangeable, to some extent. Thus, we can move some of the computational load from the encoder to the relation module without changing the performance. To test this, we simplified the encoder by removing the fully-connected layers, leaving only the convolutional layers, and complicated the relation module, by making it a more complex and expressive,
\begin{equation}
    R_{\theta}\left( Z_\phi\left(\mathbf{x}_i\right), Z_\phi\left(\mathbf{x}_j\right)\right) = H_{\theta}\left( Z_\text{conv}\left(\mathbf{x}_i\right)\oplus Z_\text{conv}\left(\mathbf{x}_j\right)\right)
\end{equation}
where the relation module $H_{\theta}$ takes a concatenation of the convolutional layers' outputs to a single output neuron and has a network architecture similar to the vanilla model's encoder's FC layers (with twice the input dimension). Rather than optimizing both the encoder and the relation module, as in the vanilla model, we optimized only the relation module. This version of the model achieved an average accuracy of $0.59\pm0.01$ (Fig. S12) comparable to that of the vanilla model, demonstrating that it is possible to move the computational load from the encoder to the relation module without paying in performance.

To conclude this section, we demonstrated two ways for carrying the computational load. Either the encoder carries most of the load by extracting the relevant feature in a manner that a simple linear relation module is sufficient for capturing the rule. Alternatively, the relation module can carry the computational load. In that case, the relation module finds a \emph{specific relation} between high-dimensional input representations, keeping the input representations fixed during problem-solving.

\subsection*{Knowledge crystallization}

Our focus so far was the ability of the networks to solve problems without any training, that is, without any accumulation of information between problems. Embedded in our model, however, is the ability to accumulate knowledge. This is because problem-solving in our model is achieved through changes in synaptic weights. This motivated us to study how solving multiple problems affects performance. In humans, the improvement of performance due to the accumulation of knowledge by training is referred to as knowledge crystallization \cite{horn_intelligencewhy_1972}. 

\begin{figure}[h!]
% \vskip 0.2in
\begin{center}
\centerline{\includegraphics[width=\linewidth]{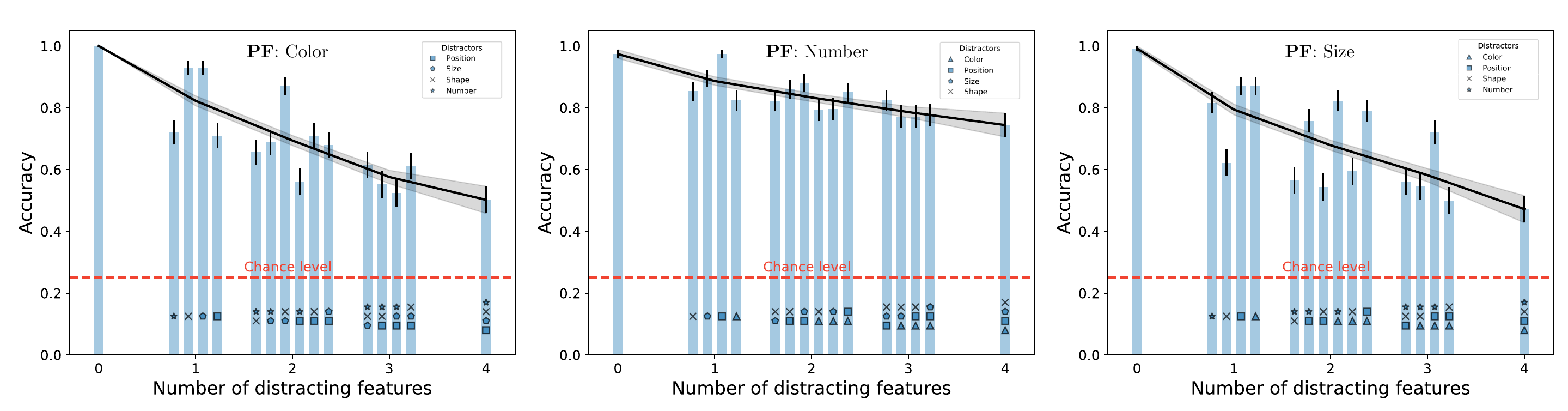}}
% \vskip -0.05in
\caption{\textbf{Knowledge crystallization.} The performance of networks that trained on 1,000 easy problems (without distracting features) of a certain predictive feature and tested on the different test conditions of that same predictive feature (test PF). Error bars correspond to 95\% CI. The black line and its shade are the average accuracy per difficulty and 95\% CI corresponding to the mean. The dashed line denotes the chance level given four choice images ($1/4$).}
\label{fig:extensive}
\end{center}
% \vskip -0.3in
\end{figure}

We first studied the extent to which the model can improve its performance on one predictive feature by practicing on that feature. Notably, in these practice sessions there was no feedback about the correct answer (in fact, the networks were exposed only to the sequences of $5$ images and not to the possible answers). We found that networks that solved 1,000 easy problems with a specific predictive feature (without resetting the weights between problems) improved their accuracy on problems with that same predictive feature to $0.74\pm0.01$ (averaged over the three predictive features, Fig. \ref{fig:extensive}), a substantial improvement from the average accuracy without prior training ($0.58\pm0.01$). Notably, the improvement was not uniform across features. While performance on Number and Size substantially improved (Size: from $0.53\pm0.01$ to $0.69\pm0.01$, Number: from $0.50\pm0.01$ to $0.84\pm0.01$), training on Color did not affect performance in Color problems ($0.70\pm0.01$ in both conditions).

Interestingly, freezing the weights of the relation module resulted in an even better performance ($0.80\pm0.01$, Fig. S13). On the other hand, the improvement was only modest when the convolutional layers' weights were frozen ($0.65\pm0.01$, Fig. S14).

This improvement in performance is analogous to knowledge crystallization. However, will training on one predictive feature improve performance when other predictive features are used? Recall that in humans, training on one task does not generalize to other tasks \cite{jaeggi_improving_2008}. Similarly, we found that while training on one predictive feature improved performance in problems with that same predictive feature, it was detrimental when the networks were tested on problems with a different predictive feature (Fig. S15). 

Will training on several predictive features improve network performance on those several trained features? To address this question, we focused on the two predictive features that exhibited improvement with training, Number and Size. We used block training and tested performance on the most difficult problems of both types. Considering the first block of training, extensively training the network with one predictive feature improves performance on that feature but not on the other feature (Fig. S16). Considering the second block, when this network is trained on the other predictive feature, the network quickly improves on that feature, but improvement on the first predictive feature quickly diminishes. when the network trains on the other feature (Fig. S16\textbf{a}, \textbf{b}). Trying to resolve this by interleaving these two predictive-feature problems in short blocks of 5 problems does not change the result and the network seems unable to simultaneously improve on two predictive features (Fig. S16\textbf{c}, \textbf{d}).

\begin{figure}[h!]
% \vskip 0.2in
\begin{center}
\centerline{\includegraphics[width=\linewidth]{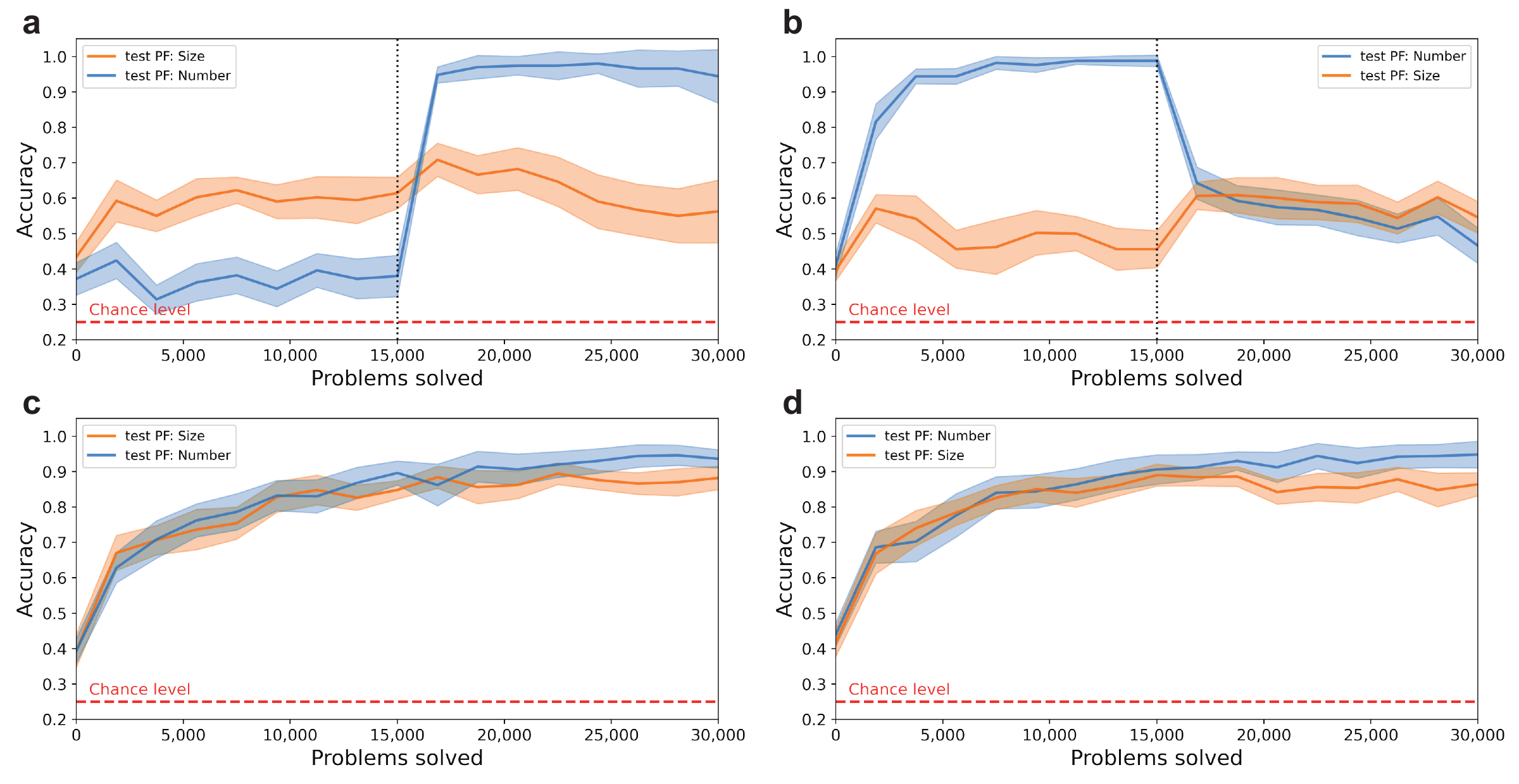}}
% \vskip -0.05in
\caption{\textbf{Interleaving effect.} Networks were trained on 15,000 problems in which the predictive feature was Size and 15,000 problems with predictive feature Number. The training problems were either presented in two large consecutive blocks (Size and then Number (\textbf{a}); Number and then Size (\textbf{b})) or interleaved at a rate of five problems per predictive feature (Size and then Number (\textbf{c}); Number and then Size (\textbf{d})). Errors correspond to 95\% CI (see Methods).}
\label{fig:interleaving}
\end{center}
% \vskip -0.3in
\end{figure}

To minimize conflict between the two features, we trained and tested the network in problems in which the competing non-predictive feature (Size or Number) was set at \emph{the same constant value} (see Methods). We found that when training was done in two long blocks, the network only improved on the trained feature (Fig. \ref{fig:interleaving}\textbf{a}, \textbf{b}). By contrast, when training was done by interleaving many short blocks of 5 problems, the network improved in both features (Fig. \ref{fig:interleaving}\textbf{c}, \textbf{d}). 

The fact that the network forgets one feature when training on the other is known in the machine learning literature as \emph{catastrophic forgetting} \cite{french_catastrophic_1999}, and indeed, interleaving has been shown to address this problem effectively \cite{robins_catastrophic_1995}. Similarly, the fact that interleaving is more effective than block training for learning is also well known in the cognitive literature as the interleaving effect \cite{kornell_learning_2008, brunmair_similarity_2019}.

\section*{Discussion}

We found that naive randomly-initialized ANNs can perform abstract reasoning that does not rely on memorization when they are optimized at test time. This result has implications both the cognitive sciences and for machine learning.

Traditionally, abstract reasoning in humans has been considered a symbolic computation -- a type of digital processing distinctly different from the analog nature of computation in ANNs \cite{turing_computable_1936, turing_computing_1950}. Recently, however, studies have shown that complex computations once attributed solely to symbolic processing can be accomplished by extensively trained ANNs \cite{krizhevsky_imagenet_2012, campbell_relational_2023}. This is especially evident with large language models, which appear capable of performing certain forms of abstract reasoning \cite{webb_emergent_2023}. Nevertheless, critics of abstract reasoning in ANNs argue that this success may be due more to sophisticated memory retrieval than to genuine abstract reasoning \cite{mccoy_embers_2023, biever_chatgpt_2023}. Our contribution is that we show that the tools used for training ANNs can also be used for exhibiting what resembles symbolic abstract reasoning without any training, hence without relying on memory recall.

A key element in our network’s ability to perform abstract reasoning tasks is the convolutional part of the encoder. We found that these random convolutional layers are instrumental in extracting features correlated with relevant latent features. Interestingly, optimization of the convolutional layers was not necessary for achieving the performance. These results resonate with human-brain studies. In humans, early visual cortex regions act as general-purpose feature extractors, sensitive to basic features like orientation and direction. It seems unlikely that this low-level feature extraction changes with every problem presented to a human participant. They may, however, change with extensive training \cite{ahissar_reverse_2004}.

Highlighting \emph{the} features that are relevant for \emph{the} particular sequence of images of a particular problem was done in the deep layers of the encoder (fully-connected layers), together with the relation module. In humans, imaging studies suggest that higher cortical regions, such as the lateral prefrontal cortex, play an important role in abstract reasoning \cite{gray_neural_2003} and rule learning \cite{mansouri_emergence_2020}. We hypothesize that these higher cortical regions perform the analog of the optimization-based computation of highlighting the relevant feature (fully-connected layers of the encoder) and identifying its regularity (the relation module).

Notably, the computations performed by the fully-connected layers of the encoder and the relation module are somewhat interchangeable. This is because either of those networks could carry the main computational load. This suggests an interesting approach for finding relations by implementing a few very simple and general (applicable to different problems) relation modules, transferring a significant computational load of finding appropriate input representations to the encoder. Furthermore, given the interchangeability of complexity in the fully-connected layers of the encoder and the relational module, the separation between the encoder and relational module in the brain analog of this computation may not exist.

Our framework naturally generalizes to explaining knowledge acquisition through problem-solving (the practice in Fig. \ref{fig:extensive} was unsupervised, with no feedback). We found that training with many short interleaved blocks was substantially more effective than training with two long blocks. This resembles a similar observation in the cognitive sciences known as the interleaving effect \cite{kornell_learning_2008, brunmair_similarity_2019}. In the cognitive sciences, two competing theories have been used to explain this effect. In one, the interleaving effect is due to the enhanced problem-identification and feature-distinction required when solving two types of problems in close proximity \cite{rohrer_interleaving_2012}. The second theory explains the interleaving effect by proposing that with interleaved training, the brain is continually engaged at retrieving the responses from memory -- a process that enhances the consolidation of those memories \cite{dunlosky_improving_2013, krug_massed_1990}. In contrast to these theories, our model has no explicit problem-identification or memory-consolidation mechanisms implemented. Rather, the interleaving effect is a manifestation of the well-known catastrophic forgetting phenomenon in machine learning \cite{french_catastrophic_1999, robins_catastrophic_1995}.

We focused in this paper on the abstract reasoning of ANNs optimized by gradient descent. These models have shown to achieve performance levels that sometimes rival or even exceed human capabilities, especially in areas like language and visual processing, key aspects of human cognition \cite{openai_gpt-4_2023, ho_denoising_2020}. However, these successes have been achieved by scaling up model size and training data, with models trained on datasets vastly larger than what human children require to learn comparable skills \cite{laurence_poverty_2001, frank_bridging_2023}. Thus, the sustainability of simply increasing network size and data volume as a path to further improvements has been doubted \cite{sun_revisiting_2017, hestness_deep_2017}. Our findings suggest a potential alternative approach, in which ANNs, by optimizing their weights at test time, exhibit computational capacity in the absence of massive datasets. This approach offers a promising direction for AI development that prioritizes efficiency over scale.

Abstract reasoning consists of several computational facets. In this work, we focused on only one of them: the identification of relationships between images in order to infer the sequence completion, also termed inductive reasoning. Inductive reasoning is needed for solving many types of intelligence tests \cite{siebers_computer_2015}. While modeling this facet, our model does not encapsulate other facets of abstract reasoning observed in humans. Specifically, our model does not incorporate working memory, limiting the regularities it can identify. It also does not explicitly perform the mapping computation required for analogical reasoning \cite{lu_probabilistic_2022}. Additionally, the model cannot solve a problem by breaking it into its sub-components \cite{carpenter_what_1990}. For example, to solve a Raven Progression Matrix, humans use the strategy of identifying common regularities in the rows and the columns. Our model was constructed only to find a regularity in a sequence. As with most ANN models, the model cannot interpret its choices. Finally, it lacks the ability to generate new images that follow the regularity it identifies. These limitations present opportunities for future research and suggest areas for improvement.

In humans, evidence suggests that abstract reasoning operates as a general computational process, analogous to a general-purpose computer that can handle any input. For example, an individual’s performance on various cognitively demanding tests tends to correlate \cite{gottfredson_general_1998}. As the tests require different prior knowledge, these correlations are taken as support for the hypothesis that a general ability, often termed general intelligence \cite{spearman_general_1904}, underlies these diverse cognitive skills. Additionally, training on a specific cognitive task usually does not improve performance on unrelated tasks \cite{jaeggi_improving_2008}. This lack of transfer suggests that human abstract reasoning is indeed general, relying on general cognitive processes rather than specific learned patterns or memorized solutions. This somewhat resembles our model.

In conclusion, our work demonstrates that ANNs can exhibit abstract reasoning abilities without reliance on memory recall, opening pathways for further exploration of abstract reasoning mechanisms in both artificial systems and humans.

\section*{Methods}

\subsection*{Code availability}

The code for this paper was written using PyTorch \cite{paszke_pytorch_2019}. The code that generates test problems and applies the model to solve them is available at \url{https://github.com/Tomer-Barak/learning-independent_abstract_reasoning}.

\subsection*{Network architectures}

The encoder ($Z(\mathbf{x})$) consisted of two main components (see Supplementary Fig. S1): three convolutional layers (kernel sizes: 2, 2, and 3; strides: all 1; padding: all 1) and five Fully-Connected (FC) layers (number of neurons: 200, 100, 50, 10, 1). Three ReLU activation functions were applied after each convolutional layer, and two Max-Pool layers (kernels: 4 and 6, strides: all 1) were applied after the second and third convolutional (+ReLU) layers. Four tanh activation functions were applied after each FC layer, except the last one, which had no activation function and remained a linear transformation.

The vanilla model's relation module consisted of a single parameter as written in equation (2). The more complex relation module written in equation (3) was implemented by a five-layer fully-connected network (number of neurons: 200, 100, 50, 10, 1). Four tanh activation functions were applied after each of this relation module's layers, except the last one, which had no activation function and remained a linear transformation.

\subsection*{Sequential visual reasoning tests}

Each image of the tests was constructed using the following five features: the number of objects in an image (possible values: 1 to 9), their shade (6 linearly distributed grayscale values), their shape (circle, triangle, square, star, hexagon), their size (6 linearly distributed values for the shapes’ enclosing circle circumference), and arrangement (a vector of grid positions that was used to place the shapes in order).

As written in the paper, the choice images' non-predictive features followed the same rules they abide by in the sequence (constant or randomly changing). The predictive feature followed the sequence rule only in the correct choice and was randomly chosen from the remaining feature values in the incorrect choices. We restricted the possibility of having a repeated choice image in the same problem.  If a repeated image was generated by chance, we generated another one to replace it.

\subsection*{Average accuracies}

In the paper, we report networks' average accuracies/performance in different experiments. For example, the vanilla model's average accuracy was $0.58\pm0.01$. These numbers were obtained (except in the knowledge crystallization section, discussed below) in the following way. For each predictive feature relevant to the experiment, we considered all its test conditions of different difficulties. There were five features, one predictive and the other four either constant or distracting, amounting to $2^4=16$ test conditions per predictive feature. We tested randomly initialized networks in $500$ problems in each test condition (each problem with a different initialized network) and obtained their success rate in that test condition. To estimate the errors, we calculated the standard error of the mean of a sample of Binomial random variables based on the success rate and the number of samples ($500$). To obtain the average accuracy of that predictive feature, we averaged the success rates over all test conditions and propagated the errors accordingly. For the total average accuracy, we averaged the accuracies of the experiment's relevant predictive features and propagated the errors.

In the knowledge crystallization section, the average accuracies (e.g., Fig. \ref{fig:extensive}) were obtained by training 50 networks in each predictive feature on 1000 easy problems (without distractors) of that predictive feature. After training, the networks solved 10 test problems in each of the 16 test conditions of a given predictive feature (results for networks that trained on one predictive feature and tested on another are shown in Fig. S15). We then calculated the average success rate of the networks in each test condition, using the standard error of the mean of Binomial random variables as errors. For the total average accuracy, we averaged across the different test conditions and all the relevant predictive features of the experiment, propagating the errors.

In the blocks versus interleaving experiments (Fig. \ref{fig:interleaving}, S16-S18), we trained 20 networks (in each of the figure panels) on 30,000 easy problems of two training predictive features. The training was either in two big blocks or interleaved into small five-problem blocks. After every 1875 training problems, we tested the networks on 25 difficult problems of the two predictive features. In Fig. S16, the easy and difficult problems were such that there were no distractors or all of the distractors. In Fig. \ref{fig:interleaving}, both the easy and difficult problems of Size had a fixed value of Number ($5$ shapes). Accordingly, the easy and difficult problems of Number had a fixed value of Size (the $5^\text{th}$ size value).

\subsection*{Correlations}

To calculate an encoder neurons' correlations with a particular feature (color, number, or size; Fig. \ref{fig:feat_corr} and S9), we generated artificial testing examples corresponding to that feature: 20 images for each of the feature's six possible values (120 examples overall) where the rest of the features' values were drawn randomly. We applied these examples to the network and recorded its neurons' activity. Based on the neurons' activity, we calculated each of the neurons' correlation with the feature values. Finally, we averaged the correlations across the layers.

To generate Fig. \ref{fig:feat_corr}\textbf{a}, we average the correlations of networks that solved the three possible predictive features with the predictive features they solved. For Fig. \ref{fig:feat_corr}\textbf{b}, we calculated the correlations with the other (non-predictive) features. In both figures, we averaged over the 16 test conditions of a given predictive feature, 50 problems per test condition, each problem solved by a different naive network. Thus, overall, the results are average over $3\times50\times6=900$ networks. The complete results of these simulations, before averaging over networks and test conditions, are shown in Fig. S8. To calculate the errors of the correlations, we estimated the standard error of the mean of the average correlations of the different networks. We propagated these errors when we averaged the correlations across test conditions and different predictive features.

To calculate the correlations (or covariance) of neurons with the sequence (Fig. \ref{fig:seq_corrs}\textbf{b-d}), we applied the sequence images to the network and recorded its neural activity. Then, we calculated for each neuron the correlation (or covariance) between its activity and the sequence order indices of the images. In Fig. \ref{fig:seq_corrs}\textbf{b-c}, we calculated the covariance with the sequence of neurons taken from the output of the encoder's $3^\text{rd}$ convolutional layer, while in Fig. \ref{fig:seq_corrs}\textbf{d}, we calculated correlations with the sequence of the FC layers' neurons. The errors in the latter case were calculated like those of the feature correlations. 

\subsection*{Figure \ref{fig:performance_vs_corr}}

To obtain the values of this plot, we considered test conditions with one distracting feature that was either Color, Number, or Size (as those are the features for which we were able to calculate networks' correlations). In total, there were $6$ such test conditions (2 for each predictive feature). For each of the test conditions, the y-axis value is the logit function ($\text{logit}(p)=\ln{\frac{p}{1-p}}$) of the success rate of $500$ problems of that test condition, each solved by a different naive network, obtained from Fig. \ref{fig:MCPC_mon}. For the errors of these values, we propagated the success rate errors through the logit function. For the x-axis values, we obtained networks' average absolute correlations with the predictive feature of the problems they solved (after solving them) and compared that with the absolute correlation of the distracting features. The values and errors were obtained from Fig. S8, and the errors were propagated through the ratio. The linear regression analysis was conducted using SciPy's \cite{virtanen_scipy_2020} linear regression function.

\subsection*{Figure \ref{fig:seq_corrs}c}

The histogram in Fig. \ref{fig:seq_corrs}\textbf{c} was obtained by considering the three predictive features (color, number, and size) and each of their 16 test conditions, 50 problems in each test condition. For each problem, we calculated the correlations between the convolutional layers' output neurons' co-variance with the sequence order and the L2 gradient norm of those neurons and averaged over the neurons. Finally, we plotted the histogram of those averages.

\section*{Data availability statement}

No datasets were generated or analyzed during the current study. The visual reasoning tests were generated in real-time by an algorithm (included in the Supplementary materials).

\bibliography{citations}

\section*{Acknowledgements}

This work was supported by the Gatsby Charitable Foundation.
Y.L. holds the David and Inez Myres Chair in Neural Computation.

\section*{Author contributions statement}

T.B. conducted the experiments, T.B. and Y.L. analyzed the results, wrote, and reviewed the manuscript.

\section*{Additional information}
\subsection*{Competing interests statement}

The authors declare no competing interests.

\end{document}